\pdfoutput=1

\documentclass[11pt]{article}

\usepackage[table]{xcolor}

\usepackage[final]{acl}

\usepackage{times}
\usepackage{latexsym}
\usepackage{booktabs}
\usepackage{multirow}
\usepackage{graphicx}
\usepackage{amsfonts}
\usepackage{enumitem}
\usepackage{xcolor}

\usepackage[T1]{fontenc}

\usepackage[utf8]{inputenc}

\usepackage{microtype}

\usepackage{inconsolata}

\usepackage{graphicx}

\usepackage{tabularx}

\usepackage{amsmath}

\title{CHEER-Ekman: Fine-grained Embodied Emotion Classification}

\author{Phan Anh Duong, Cat Luong, Divyesh Bommana, Tianyu Jiang \\
  University of Cincinnati \\
  \texttt{\{duongap, luongcn, bommandh\}@mail.uc.edu, tianyu.jiang@uc.edu} \\}

\begin{document}
\maketitle
\begin{abstract}
Emotions manifest through physical experiences and bodily reactions, yet identifying such embodied emotions in text remains understudied. We present an embodied emotion classification dataset, CHEER-Ekman,\footnote{\url{https://github.com/menamerai/cheer-ekman}} extending the existing binary embodied emotion dataset with Ekman's six basic emotion categories. Using automatic best-worst scaling with large language models, we achieve performance superior to supervised approaches on our new dataset. Our investigation reveals that simplified prompting instructions and chain-of-thought reasoning significantly improve emotion recognition accuracy, enabling smaller models to achieve competitive performance with larger ones.
\end{abstract}

\section{Introduction}

Emotions are not merely abstract mental states; they are deeply intertwined with somatic experiences. When we feel joy, our faces light up with smiles; when we are scared, our hearts race and our hands tremble. These physical reactions are more than just side effects---they are part of how we experience and express emotions. This concept, known as embodied emotion, suggests that our bodies play a key role in how we feel, perceive, and understand emotions~\citep{lakoff_philosophy_1999, niedenthal_embodying_2007}. In natural language, these connections surface as descriptions of physiological reactions (e.g., ``my stomach churned in disgust'') or unintentional physical actions (e.g., ``she stomped her feet in frustration'')---phenomena termed \textit{embodied emotions}. Recognizing such expressions is pivotal for understanding implicit emotional cues in narratives. While recent advances in NLP have focused more on explicit emotion classification~\citep{mohammad-etal-2018-semeval} or sentiment analysis~\citep{rosenthal-etal-2017-semeval}, the subtler task of identifying embodied emotions remains less explored, despite its psychological grounding and practical relevance.

\begin{figure}[t]
    \centering
    \includegraphics[width=0.95\columnwidth]{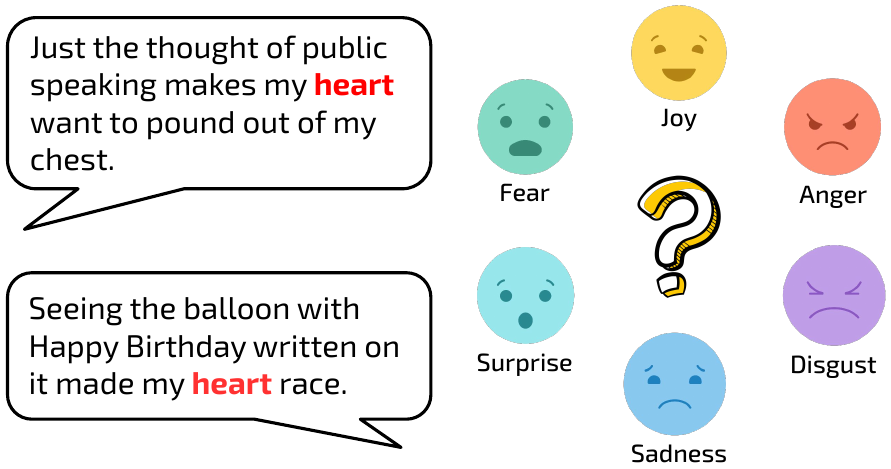}
    \caption{Illustration of embodied emotions classified into six categories.}
    \label{fig:intro}
\end{figure}

The CHEER dataset \citep{zhuang-etal-2024-heart} filled a gap in this field by providing a collection of sentences where body parts are used to express emotions. The dataset includes 7,300 human-annotated sentences containing body part references, proposing a binary classification task which we will refer to as ``embodied emotion detection.'' However, one limitation of this work is that it does not distinguish between different types of emotions---for instance, whether a racing heart signals fear or excitement. The framework of \citeposs{ekman_are_1992} basic emotions offers a potential solution to this limitation. By linking embodied expressions to these specific emotions, we can build systems that better understand human emotional experiences.

To achieve this goal, we extend the CHEER dataset by annotating all its 1,350 positive samples with six Ekman emotion labels (\textit{Joy}, \textit{Sadness}, \textit{Anger}, \textit{Disgust}, \textit{Fear}, and \textit{Surprise}), creating a new dataset, \textbf{CHEER-Ekman}, as illustrated in Figure~\ref{fig:intro}. For clarity, we refer to the novel classification task produced by this dataset as ``embodied emotion \textit{classification},'' as compared to the binary task discussed above. We adopt the automatic best-worst scaling (BWS) technique~\citep{kiritchenko-mohammad-2017-best, bagdon-etal-2024-expert} with large language models (LLMs) to tackle the task. Our experiments show that using Llama 3.1 8B with BWS significantly outperforms zero-shot prompting. The best BWS experiment achieved a 50.6 F1-score, surpassing supervised BERT (49.6) and beating zero-shot approaches by around 20 points. Building on \citeposs{zhuang-etal-2024-heart} investigation of LLMs' capability for embodied emotion detection, we further explore prompting techniques that enhanced the detection task. Our experimental results reveal that LLMs can make better recognition when instructions are rephrased in plain, easily understood language, which boosts F1 by nearly 30 points compared to technical definitions. Moreover, chain-of-thought reasoning enables an 8B parameter model to nearly match a 70B model, closing the performance gap to within 7 F1 points. In summary, our contributions are three-fold:

\begin{enumerate}[topsep=2mm]
\item We present CHEER-Ekman, an extension of the CHEER dataset that enriches embodied emotion expressions with fine-grained Ekman emotion labels, addressing a critical gap in understanding how specific emotions manifest through bodily expressions. Our dataset is available at: \url{https://github.com/menamerai/cheer-ekman}.
\vspace{-1mm}

\item We demonstrate that the automatic best-worst scaling technique enables LLMs to perform emotion classification without any task-specific training, achieving performance that exceeds supervised approaches.
\vspace{-1mm}

\item We reveal that counterintuitively, simplified everyday language in prompts dramatically outperforms technical definitions for embodied emotion tasks, and that structured reasoning through chain-of-thought can allow smaller language models to perform at a level closer to larger language models on our task.

\end{enumerate}

\section{Related Work}
Emotion recognition in natural language processing has been extensively studied, with researchers focusing more on explicit emotion using datasets like SemEval~\citep{strapparava-mihalcea-2007-semeval, mohammad-etal-2018-semeval} and GoEmotions~\citep{demszky2020goemotions}.  Recent research has expanded to explore nuanced aspects of emotion expression~\citep{li2021learning}, including emotion intensity prediction~\citep{mohammad2018obtaining, bagdon-etal-2024-expert} and the detection of subtle emotional cues in dialogues~\citep{poria2018meld, ghosal-etal-2020-cosmic, li2022knowledge}. The advent of large language models (LLMs) has catalyzed significant advances in emotion understanding capabilities~\citep{lee-etal-2024-pouring, sabour-etal-2024-emobench, zhao-etal-2024-matter, Liu-etal-2024-EmoLLMs}.

While these works contribute to a deeper understanding of emotion detection in text, the embodied nature of emotions---how physical sensations and actions encode affective states---has received comparatively less attention. The concept of embodied emotion is rooted in cognitive science, particularly in the works of \citet{lakoff_philosophy_1999} and \citet{niedenthal_embodying_2007}, which suggest that emotional experiences are closely tied to bodily states and actions. Despite the psychological grounding of embodied emotions, computational approaches to capturing them in text remain limited. \citet{zhuang-etal-2024-heart} introduced the CHEER dataset, which provides a collection of sentences where body parts are explicitly used to express emotions. Our work extends theirs by incorporating Ekman’s six basic emotions into embodied emotion recognition, offering a more fine-grained classification system.

Another relevant line of work is the study of emotion taxonomy. Although recent advances in psychology have offered newer granular categories of emotions such as 27 emotions by \citet{Cowen-27-emotion}, which has been adopted in both textual emotion datasets~\citep{demszky2020goemotions} and visual emotion dataset~\citep{kosti2019context}, we follow the vast majority of existing emotion datasets~\citep{strapparava-mihalcea-2007-semeval,mohammad-etal-2018-semeval,poria2018meld} by utilizing the six basic emotions (\textit{Joy}, \textit{Sadness}, \textit{Anger}, \textit{Disgust}, \textit{Fear}, and \textit{Surprise}) proposed by \citet{ekman_are_1992}, which remain foundational due to their universality and simplicity. Future research may explore integrating alternative taxonomies into embodied emotion classification to enhance both granularity and coverage. 

\section{Methods}

Our methodological approach comprises three key components that build upon and extend the work of~\citet{zhuang-etal-2024-heart}. First, we explore prompting strategies to enhance LLMs' capability to detect embodied emotions. Second, we introduce CHEER-Ekman, a refinement of the original CHEER dataset that adds fine-grained emotion labels. Finally, we adopt the BWS framework for emotion classification that leverages comparative judgments to improve classification accuracy.

\subsection{Prompting LLMs for Embodied Emotion Detection}
To address the gap in prompt design within the \citeposs{zhuang-etal-2024-heart} framework, we first sought to enhance the embodied emotion detection task using state-of-the-art LLMs and explore the impact of prompt engineering on performance. Our approach centers on two key strategies: prompt simplification to mitigate linguistic complexity and chain-of-thought (CoT) prompting.

\paragraph{Prompt Simplification.} 
We investigated the effects of linguistic and domain complexity by conducting experiments with the base Llama-3.1~\citep{grattafiori2024llama3herdmodels} and the recently released DeepSeek-R1 distilled version~\citep{deepseekai_2025_deepseekr1}. Specifically, we compared two prompts: the base prompt used in \citet{zhuang-etal-2024-heart} and a simplified prompt, which reduces syntactic and lexical complexity to minimize potential comprehension barriers for LLMs.

\paragraph{Chain-of-Thought Prompting.} 
We further explored eliciting reasoning from the model by implementing chain-of-thought (CoT) prompting. Based on \citeposs{zhuang-etal-2024-heart} annotation criteria for embodied emotion detection, we developed three CoT variants: a 2-step variant that evaluates emotional causation and purposeless expression, a 3-step variant that adds body part identification, and a simplified 2-step variant with reduced linguistic complexity. These variants allowed us to examine how explicit causal reasoning affects both the model's emotion detection performance and its understanding of body-emotion relationships.

\subsection{Dataset Creation}

While embodied emotion detection identifies emotional expressions through bodily movements, understanding the specific emotions conveyed requires more fine-grained annotation. To address this need, we propose \textbf{CHEER-Ekman}, a refined dataset extending the original CHEER corpus \citep{zhuang-etal-2024-heart} by annotating its 1,350 positive embodied emotion instances with \citeposs{ekman_are_1992} six basic emotions (\textit{Joy}, \textit{Sadness}, \textit{Anger}, \textit{Disgust}, \textit{Fear}, and \textit{Surprise}). Our adoption of Ekman's basic emotions taxonomy balances granularity with practical considerations, as recent research by \citet{Liu-etal-2024-EmoLLMs} demonstrates that finer-grained emotion taxonomies often face sparsity issues, even in bigger datasets like GoEmotions \cite{demszky2020goemotions}. 

\begin{figure}[t]
    \centering
    \includegraphics[width=0.75\linewidth]{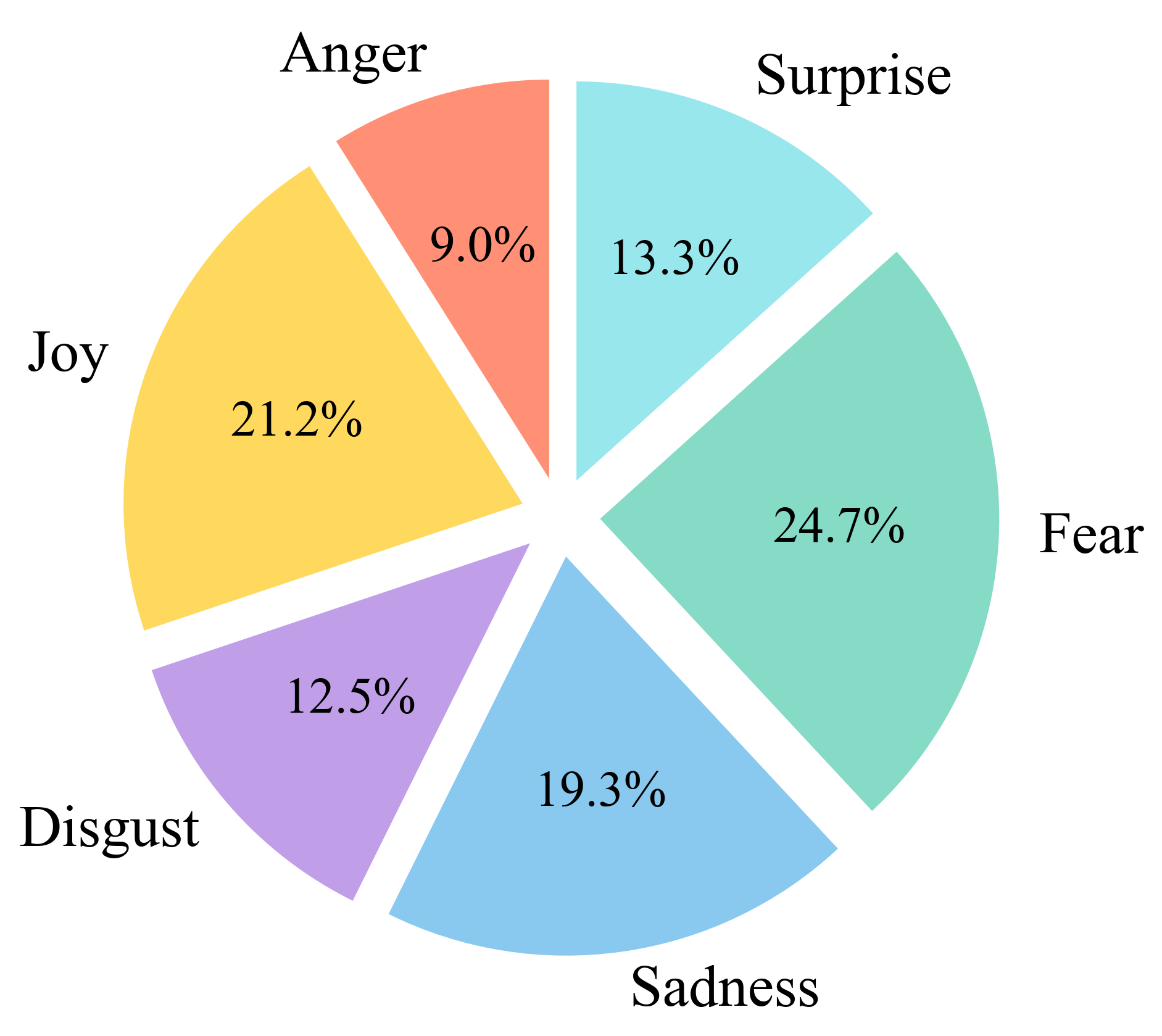}
    \caption{CHEER-Ekman dataset distribution of emotions.}
    \label{fig:emotion-dis}
\end{figure}

This approach also maintains consistency with \citeposs{zhuang-etal-2024-heart} methodology, which utilized emotion-associated adverbs derived from these basic emotions for weak supervision. We have also elected not to include the weakly labeled positive samples from the CHEER dataset, prioritizing our final dataset quality and reliability over quantity.

We recruited two annotators to label the 1,350 embodied emotion sentences from the original CHEER dataset. For each sentence, we provide the annotators with the sentence, the relevant body part, and up to three preceding sentences for context. We then ask the annotators to select one of the six emotions that best match the physical experience described through the body part. The pairwise inter-annotator agreement by Cohen's Kappa is 0.64, indicating good agreement. Finally, the annotators adjudicated their disagreements to produce the final gold labels. We show some example sentences with their annotated emotions in Table~\ref{tab:emo-labels}.

Figure~\ref{fig:emotion-dis} illustrates the emotion distribution of sentences in our newly constructed CHEER-Ekman dataset. Specifically, \textit{Fear} is the most prevalent emotion (24.7\%), followed by \textit{Joy} (21.2\%), \textit{Sadness} (19.3\%), \textit{Surprise} (13.3\%), and \textit{Disgust} (12.5\%). \textit{Anger} appeared last at 9.0\%.

\begin{table}[t]
\centering
\resizebox{\columnwidth}{!}{%
\begin{tabular}{ll}
\toprule
\textbf{Emotion} & \textbf{Sentence} \\ \midrule
Joy & ... watched the fireflies with a loving look on his \textbf{face}. \\
Sadness & ... frowning and scuffing his \textbf{feet} along the floor. \\
Fear & Marty nervously runs his \textbf{fingers} through his hair... \\
Anger & ... makes me want to hit my \textbf{head} against the wall. \\
Disgust & Dean snorted incredulously, shaking his \textbf{head} in disbelief. \\
Surprise & ... my \textbf{eyes} almost fell out of my head. \\ \bottomrule
\end{tabular}%
}
\caption{Examples in our CHEER-Ekman datasets.}
\label{tab:emo-labels}
\end{table}

\subsection{BWS for Emotion Classification}
\label{sec:bws_emo_class}

To address the fine-grained emotion classification task proposed with CHEER-Ekman, we first tested zero-shot LLM prompting. However, the model often fails to adhere to the instructions and is prone to erroneous behaviors, leading to incorrect outputs. Inspired by recent work on automatic emotion intensity annotation using LLMs~\citep{bagdon-etal-2024-expert}, we adopted the same best-worst scaling technique. The methodology involves presenting the LLM with tuples of four different sentences, instructing it to identify the body part instances that most and least represent a specific Ekman emotion. Then, the equation 
$\frac{\#Best - \#Worst}{\#Total}$ is used to calculate a score per sentence per emotion. Finally, we choose the emotion that receives the highest score to be the prediction for the sentence.

To examine how the number of comparisons affects classification accuracy, we tested a broader range of tuple counts, from $2N$ to $72N$, increasing by 50\% at each step of expansion (where $N$ is the number of instances to be classified). While \citet{bagdon-etal-2024-expert} found that more tuples improved accuracy in their experiments up to 12N, we expanded this investigation to 72N to further explore performance gain behaviors from scaling comparative rounds.

\section{Evaluation}

We conduct experiments to tackle both the embodied emotion detection and emotion classification tasks. The embodied emotion detection CHEER dataset contains 7,300 sentences, and our CHEER-Ekman dataset contains 1,350 sentences. We explored various strategies and models, and reported their F1-scores on both datasets.

\subsection{Embodied Emotion Detection}

\begin{table}[t]
\centering
\resizebox{\columnwidth}{!}{%
\begin{tabular}{lccccccc}
\toprule
\multirow{2}{*}[-0.3em]{\textbf{Model}} & \multirow{2}{*}[-0.3em]{\begin{tabular}[c]{@{}c@{}}\textbf{Macro}\\ \textbf{F1}\end{tabular}} & \multicolumn{3}{c}{\textbf{EE}} & \multicolumn{3}{c}{\textbf{Neutral}} \\
\cmidrule(lr){3-5} \cmidrule(lr){6-8}
 &  & \textbf{Pre} & \textbf{Rec} & \textbf{F1} & \textbf{Pre} & \textbf{Rec} & \textbf{F1} \\ \midrule

Llama$_{\text{base}}$ & 37.2 & 21.5 & 99.6 & 35.3 & 99.6 & 24.2 & 39.0 \\
Llama$_{\text{simple}}$ & 66.7 & 37.5 & 89.3 & 52.8 & 96.9 & 68.9 & 80.6 \\ \midrule
DeepSeek$_{\text{base}}$ & 32.6 & 20.3 & 99.4 & 33.7 & 99.3 & 18.7 & 31.5 \\
DeepSeek$_{\text{simple}}$ & 74.2 & 51.2 & 69.3 & 58.9 & 93.1 & 86.2 & 89.5 \\ \midrule
GPT 3.5$_{\text{base}}$ & 70.2 & 44.0 & 68.3 & 53.5 & 92.5 & 81.9 & 86.9 \\
BERT & 83.5 & 73.2 & 72.1 & 72.6 & 94.2 & 94.5 & 94.4 \\ \bottomrule
\end{tabular}%
}
\caption{Results comparison for Embodied Emotion Detection. Llama: Llama-3.1-70B. DeepSeek: DeepSeek-R1-Distilled-Llama-70B. GPT 3.5 and fine-tuned BERT numbers are  from~\citet{zhuang-etal-2024-heart}. The \textit{base} and \textit{simple} subscripts indicate the type of prompts, which can be found in Table~\ref{tab:zs-fs-prompts}.}
\label{tab:eec-class}
\end{table}

\paragraph{Simple Prompting Analysis.} 
To obtain the binary classification results, we directly compare the logit probabilities of ``\textit{True}'' and ``\textit{False}'' tokens instead of using text generation. This approach ensures deterministic outputs by avoiding the randomness inherent in sampling-based decoding methods, while also preventing potential output format violations that can occur during free-form generation. Table~\ref{tab:eec-class} shows that simplified prompts led to substantial performance improvements for the 70B parameter models. The F1-score increased by 29.5 points for Llama-3.1-70B, and by 41.6 points for DeepSeek-R1-70B (distilled on Llama), surpassing GPT 3.5 results reported in~\citet{zhuang-etal-2024-heart}.

\begin{table}[t]
\centering
\resizebox{\columnwidth}{!}{%
\begin{tabular}{lccccccc}
\toprule
\multirow{2}{*}[-0.3em]{\textbf{Model}} & \multirow{2}{*}[-0.3em]{\begin{tabular}[c]{@{}c@{}}\textbf{Macro}\\ \textbf{F1}\end{tabular}} & \multicolumn{3}{c}{\textbf{EE}} & \multicolumn{3}{c}{\textbf{Neutral}} \\
 \cmidrule(lr){3-5} \cmidrule(lr){6-8}
 &  & \textbf{Pre} & \textbf{Rec} & \textbf{F1} & \textbf{Pre} & \textbf{Rec} & \textbf{F1} \\ \midrule
Llama$_{\text{2-step}}$ & 53.4 & 26.2 & 80.8 & 39.6 & 93.0 & 52.7 & 67.2 \\
Llama$_{\text{3-step}}$ & 54.8 & 24.9 & 53.4 & 34.0 & 87.3 & 66.5 & 75.5 \\
Llama$_{\text{2-step-simple}}$ & 60.1 & 31.5 & 44.5 & 36.9 & 87.4 & 79.8 & 83.4 \\ \midrule
DeepSeek$_{\text{2-step}}$ & 52.2 & 26.4 & 90.8 & 40.9 & 96.1 & 47.3 & 63.4 \\
DeepSeek$_{\text{3-step}}$ & 57.4 & 27.9 & 62.0 & 38.5 & 89.4 & 66.7 & 76.4 \\
DeepSeek$_{\text{2-step-simple}}$ & 67.5 & 40.1 & 65.2 & 49.7 & 91.7 & 79.8 & 85.3 \\ \bottomrule
\end{tabular}%
}
\caption{CoT results for Embodied Emotion Detection. Llama: Llama-3.1-8B. DeepSeek: DeepSeek-R1-Distilled-Llama-8B. The \textit{2-step}, \textit{2-step-simple}, and \textit{3-step} subscripts indicate the type of prompt accompanying the model in that run. Prompt details are in Table~\ref{tab:cot-prompts}.}
\label{tab:eed-cot}
\end{table}

\paragraph{Chain-of-Thought (CoT) Analysis.} 
Table~\ref{tab:eed-cot} shows that CoT prompting enhanced performance to competitive levels with larger models in the experiments of Table~\ref{tab:eec-class}, particularly benefiting distilled reasoning models like DeepSeek-R1-8B (distilled on Llama). The DeepSeek 8B model using simple 2-step prompts (DeepSeek$_{\text{2-step-simple}}$) achieved results within 6.7 F1-points of its larger 70B counterpart (DeepSeek$_{\text{simple}}$) and 2.7 F1-points of GPT 3.5. Deeper reasoning processes proved more effective, with 3-step CoT consistently outperforming 2-step variants across both models. Finally, simplified prompting substantially improved CoT performance, yielding F1-score increases of 6.7 (Llama$_{\text{2-step-simple}}$ vs. Llama$_{\text{2-step}}$) and 15.3 (DeepSeek$_{\text{2-step-simple}}$ vs. DeepSeek$_{\text{2-step}}$).

\paragraph{Error Analysis.} To investigate model failures in the zero-shot experiments using the \textit{simple} prompt setting with Llama-3.1-70B, we analyzed incorrect predictions. We found a pronounced false-positive bias, accounting for 93.3\% of all errors. A manual inspection of 100 false-positive cases revealed three main patterns. First, 17\% of cases involved referenced body parts that were present in the experience or expression without acting, as in ``\textit{tears falling down the \textbf{face}}.'' Second, 42\% of errors stemmed from body parts performing functional or physiological roles within emotional contexts, such as eyes closing when ``\textit{blackness crept across his \textbf{eyes}},'' a natural physiological reaction associated with the character passing away within the context. Finally, 41\% of errors involved metaphorical or idiomatic expressions. These included cases where emotional embodiment was implied but not explicitly stated (``\textit{I couldn't believe my \textbf{eyes},}'' implying widened eyes in surprise but not explicitly describing this action), expressions symbolically referring to emotional states without literal physical embodiment (``\textit{a straw that broke my \textbf{back}}''), or purely locational expressions involving body parts without any action (``\textit{thoughts racing through my \textbf{head}}''). These nuanced distinctions highlight the model's challenges in accurately interpreting metaphorical, symbolic, and non-embodied references.

\subsection{Embodied Emotion Classification}
\label{sec:ee_classification}

\begin{table}[t]
\centering
\resizebox{\columnwidth}{!}{%
\begin{tabular}{lccccccc}
\toprule
\textbf{Model} & \textbf{F1} & \textbf{F1-J} & \textbf{F1-Sa} & \textbf{F1-F} & \textbf{F1-A} & \textbf{F1-D} & \textbf{F1-Su} \\ \midrule
Llama & 31.6 & 39.4 & 43.6 & 26.6 & 32.2 & 19.1 & 28.5 \\
DeepSeek & 28.4 & 43.3 & 35.7 & 33.1 & 23.1 & 14.8 & 20.2 \\ \midrule
BWS$_{4N}$ & 41.8 & 62.3 & 57.7 & 37.9 & 28.1 & 30.1 & 33.9 \\
BWS$_{12N}$ & 44.6 & 67.1 & 59.2 & 44.3 & 38.6 & 19.0 & 39.3 \\
\rowcolor{gray!20} BWS$_{36N}$ & \textbf{50.6} & 66.7 & \textbf{64.7} & 48.0 & \textbf{53.2} & 22.0 & \textbf{48.9} \\
BWS$_{48N}$ & 49.8 & 68.0 & 62.8 & 48.2 & 51.3 & 24.8 & 43.8 \\
BWS$_{72N}$ & 49.5 & \textbf{68.5} & 64.5 & 46.2 & 51.6 & 20.5 & 45.6 \\ \midrule
BERT & 49.6 & 68.2 & 57.5 & \textbf{50.1} & 30.2 & \textbf{56.1} & 35.7 \\ \bottomrule
\end{tabular}%
}
\caption{Results for Emotion Classification. Llama: Llama-3.1-8B. DeepSeek: DeepSeek-R1-Distilled-Llama-8B. BWS: Automatic BWS with Llama-3.1-8B. The first column F1 is the macro-averaged score, followed by F1-score F1-x, where J - \textit{Joy}, Sa - \textit{Sad}, F - \textit{Fear}, A - \textit{Anger}, D - \textit{Disgust}, and Su - \textit{Surprise}.}
\label{tab:ec-class}
\end{table}

Our experimental results demonstrate notable performance disparities across prompting strategies and model architectures. We use the Llama-3.1-8B as the LLM interpreter for best-worst scaling (BWS). In Table~\ref{tab:ec-class}, the first section shows the performance of zero-shot large language models, including Llama-3.1-8B and DeepSeek-R1-8B. The second section shows BWS results with different numbers of tuples. And the last section shows a fine-tuned BERT model for comparison (details in Appendix~\ref{app:bert}). We see that BWS exhibits superior performance even with smaller tuple configurations ($4N$), exceeding Llama-3.1-8B by 10.2 points. Performance improves consistently as the number of tuples increases from $2N$ to $36N$ (40.2 to 50.6), suggesting enhanced classification from expanded pairwise comparisons. We hypothesize that this expansion helps the model better weigh emotional significance in text, improving classification accuracy. Notably, the best result comes from the expansion to $36N$ tuples, with the F1-score beating the supervised method BERT by 1 point. Our experimental results show that as we keep increasing $N$, the performance will reach a plateau as evidenced by $48N$ and $72N$ (see Appendix~\ref{app:bws_n}). 

\paragraph{Error Analysis.} 
To better understand model limitations, we conducted a qualitative error analysis on the misclassified cases. We identified several consistent failure modes, including the model's difficulty in interpreting emotionally complex inputs or making reliable distinctions between closely related emotional states. In one case, the model predicted \textit{Joy}, even though the embodied expression \textit{``Ryan ducks his \textbf{head} down to his notebook''} signaled \textit{Fear}. This misclassification likely resulted from the influence of nearby positive context, such as \textit{``Brendon waves and smiles''}, which distracted the model from the emotion-relevant phrase. In another case, the vivid scene \textit{``the age-old rock tradition of holding up lighters spread across the 28{,}000 person deep crowd \dots{}lighting up the entire audience \dots{}the hair on my \textbf{arms} started to raise''} was misclassified as \textit{Joy}, despite strong physiological cues such as raised arm hair that more closely reflect \textit{Surprise}. This highlights the model’s tendency to prioritize surface-level celebratory language over conflicting embodied cues.

\section{Conclusion}

This work advances embodied emotion recognition through three main contributions. First, we created a new dataset called CHEER-Ekman by extending the CHEER dataset with Ekman emotion labels to better understand the connection between bodily expressions and emotional states. Second, we demonstrated that best-worst scaling outperforms both prompted LLMs and fine-tuned BERT, showing the potential for emotion classification without task-specific training. Finally, we found that simplified language and chain-of-thought reasoning significantly improve LLM performance in embodied emotion detection, enabling smaller models to achieve competitive results.

\section*{Limitations}
While our approach demonstrates a promising advancement in embodied emotion detection using LLMs and the best-worst scaling technique, several limitations warrant consideration.

First, a key observation in our embodied emotion detection task was that simplifying prompts significantly improved model performance. While these findings may suggest enhanced efficiency through linguistic streamlining, they simultaneously introduce concerns about potential overfitting to these simplified phrasings. Simplified prompts may inadvertently prioritize more explicit expressions of embodied emotion over subtler or more figurative language, meaning the models might learn to recognize patterns specific to the prompt structure rather than generalizing to a wide variety of natural language expressions.

Second, the CHEER-Ekman dataset is relatively small, consisting of only 1,350 sentences. This limited size stems from our decision to annotate only the sentences already identified as containing embodied emotions in the original CHEER dataset. This selective annotation was intended to efficiently focus our efforts on instances most relevant to embodied emotion, but it may introduce a bias towards positive examples.

Finally, when addressing emotion classification via best-worst scaling, the scalability and computational overhead of this methodology present challenges: while higher tuple quantities lead to higher accuracy, they also impose significant computational costs. Due to time and computational constraints, we utilize a smaller model, which may lead to suboptimal results for higher-order tuples. Additionally, the limited context window prevents us from effectively implementing a few-shot setting, further impacting performance in scenarios requiring extended context understanding.

\section*{Acknowledgments}
We thank the CincyNLP group for their constructive comments. We also thank the anonymous ACL reviewers for their valuable suggestions and feedback.

\newpage

\bibliography{custom}

\newpage

\appendix

\section{LLM Prompt for Embodied Emotion Detection \& Embodied Emotion Classification}
\label{app:eed_prompt}

Table~\ref{tab:zs-fs-prompts}-\ref{tab:emo-prompts} presents the complete prompt templates employed in our experimental methodology. Table~\ref{tab:zs-fs-prompts} details the prompts used for zero-shot embodied emotion detection experiments. The \textit{Base} prompt closely replicates the methodology of \citet{zhuang-etal-2024-heart}, with the sole modification being the use of ``True'' and ``False'' as decision tokens rather than ``Yes'' and ``No''. The \textit{Simple} prompt maintains the fundamental logic while employing more straightforward language and structure to reduce cognitive complexity.

Table~\ref{tab:cot-prompts} presents our chain-of-thought (CoT) prompt variants. The \textit{2-step} implementation incorporates the dual criteria from the Base prompt as explicit reasoning steps. The \textit{3-step} variant augments this with an initial body movement identification phase, designed to establish concrete context and facilitate more comprehensive reasoning. The \textit{2-step simple} variant examines the effectiveness of linguistic simplification within the CoT framework.

Table~\ref{tab:emo-prompts} outlines the large language model (LLM) prompt utilized for emotion classification experiments with Llama-3.1-8B and DeepSeek-R1-8B models.

Throughout our prompt templates, we employ the following placeholder semantics:

\begin{itemize}
    \item ``\texttt{<sentence|>}'' denotes the target sentence containing the body part for evaluation.
    \vspace{-1.5mm}
    \item ``\texttt{<bdypart|>}'' indicates the specific body part instance within the sentence.
    \vspace{-1.5mm}
    \item ``\texttt{<preceed|>}'' represents up to three preceding context sentences, when available.
\end{itemize}

This placeholder convention remains consistent across all experimental tasks presented in this research.

\section{Best-Worst Scaling}
\label{app:bws_n}
\noindent\textbf{Experiment Setup.} 
Best-worst scaling (BWS) is a comparative annotation method where annotators select the best and worst items from a given set, typically a 4-tuple. This approach efficiently derives pairwise comparisons, as selecting the best and worst items provides information about most item relationships within the set. A single annotation with best-worst scaling is equivalent to an annotation with 6 pairwise comparisons. Hence, using BWS allows for fewer inferences with the same result. This is particularly beneficial for identifying emotion intensity or emotion classification.

These 4-tuples are assembled from the test data and then are presented to LLMs using the prompt in Table~\ref{tab:emo-bws-prompts}. The model then picks one instance that most represents and one instance that least represents some property (in our case, this would be one of the six \citeposs{ekman_are_1992} emotions).

Once multiple 4-tuples are annotated, a simple counting procedure generates numerical scores, allowing items to be ordered according to their relevance to the given property. The score is calculated using the formula $\frac{\#Best - \#Worst}{\#Overall}$, where the $\#Best$ and $\#Worst$ represent the number of times a sentence is ranked Best or Worst, respectively; and $\#Overall$ denotes the number of occurrences of the sentence across all 4-tuples. This approach captures a continuous measure, reflecting the relative intensity of a sentence within the given category.

With this method, the LLMs can perform accurate annotations. The results from the annotations will be calculated to get a BWS intensity score across 6 Ekman emotions. The emotion with the highest intensity score will be chosen as the predicted label. This classification process can be represented by the following expression:
\[
\hat{e} = \arg\max_{e_i \in E} S(e_i)
\]
where $e_i$ corresponds to each emotion, $E$ is the set of all emotions, and $S(e_i)$ is the intensity score w.r.t. to such emotion. The resulting predictions are compared with the labels from the CHEER-Ekman dataset to assess performance using several metrics, with a particular focus on the F1-score. The approach of increasing the number of tuples to enhance performance was proposed by \citet{bagdon-etal-2024-expert}.

Along with the embodied emotion classification prompt, we also incorporate two additional placeholder semantics:

\begin{itemize}
    \item ``\texttt{<textid|>}'' denotes the unique instance ID from the dataset. This id helps the model easily pick out its answer from the sentence tuple when inferencing.
    \item ``\texttt{<emo|>}'' denotes the specific emotion required for ranking.
\end{itemize}

This placeholder convention remains consistent across all experimental tasks presented in this research.

\begin{figure*}[h]
    \centering
\includegraphics[width=0.95\linewidth]{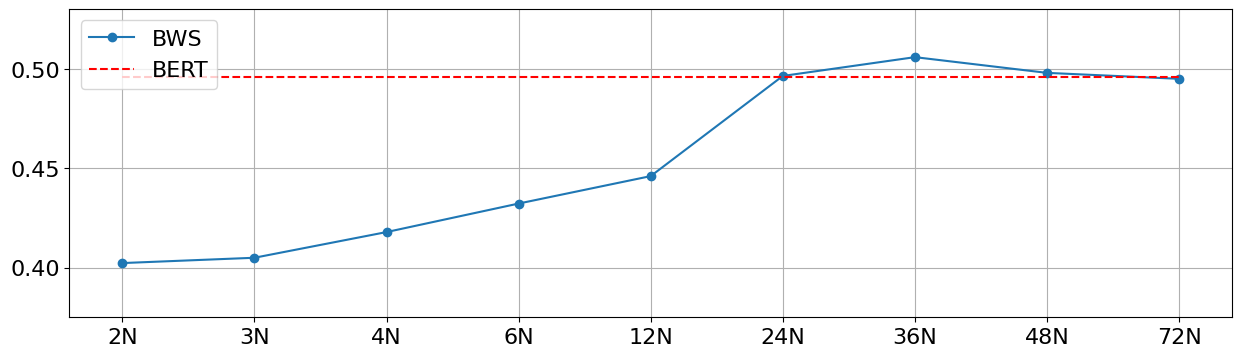}
    \caption{F1-score trends for BWS when increasing the number of tuples from 2N to 72N (where N is the total number of instances to be classified). BERT’s performance is shown as a reference baseline.}
    \label{fig:bws-scale}
\end{figure*}

\noindent\textbf{Performance Plateau.} 
Figure~\ref{fig:bws-scale} illustrates the relationship between tuple count and F1-score performance in our BWS experiments. The results demonstrate significant performance improvements up to $24N$ tuples, reaching optimal performance at $36N$. Beyond this threshold, performance degradation is observed, with decreased F1-scores at both $48N$ and $72N$ tuple configurations. This pattern suggests a clear upper bound for effective tuple scaling in BWS implementations.

\section{BERT Experiment Setup}
\label{app:bert}
For the embodied emotion classification task discussed in Section~\ref{sec:bws_emo_class}, in addition to BWS, we fine-tuned BERT as a reference benchmark \citep{devlin-etal-2019-bert}. Inputs were constructed by concatenating the preceding context, main text, and referenced body part. We set the maximum sequence length to 512 and use a batch size of 16. The model was trained with the AdamW optimizer, a learning rate of 2e-5, and evaluated over 15 training epochs. Cross-entropy loss was used as the objective, with tokenization performed using truncation and padding. All runs were conducted with 5 seed values starting from 41 to 45 for reproducibility, and final results were averaged across these five runs.

\section{Data and Results Analysis}
\label{app:data_and_results_analysis}

\begin{figure*}[h]
    \centering
    \includegraphics[width=0.95\linewidth]{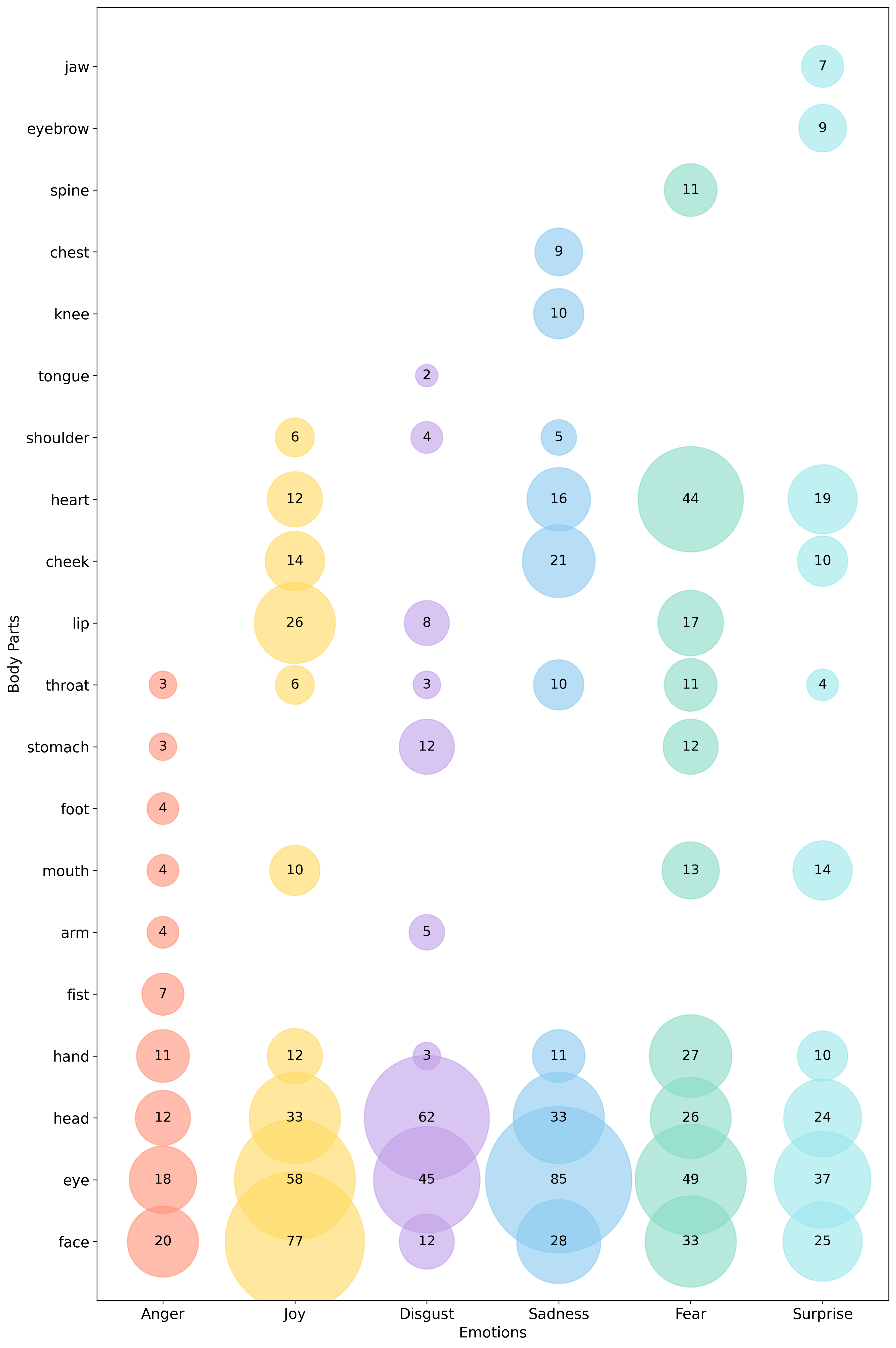}
    \caption{Frequency of top 10 body parts for each emotion.}
    \label{fig:bodypart_distributions}
\end{figure*}

\begin{figure*}[h]
    \centering
    \includegraphics[width=0.95\linewidth]{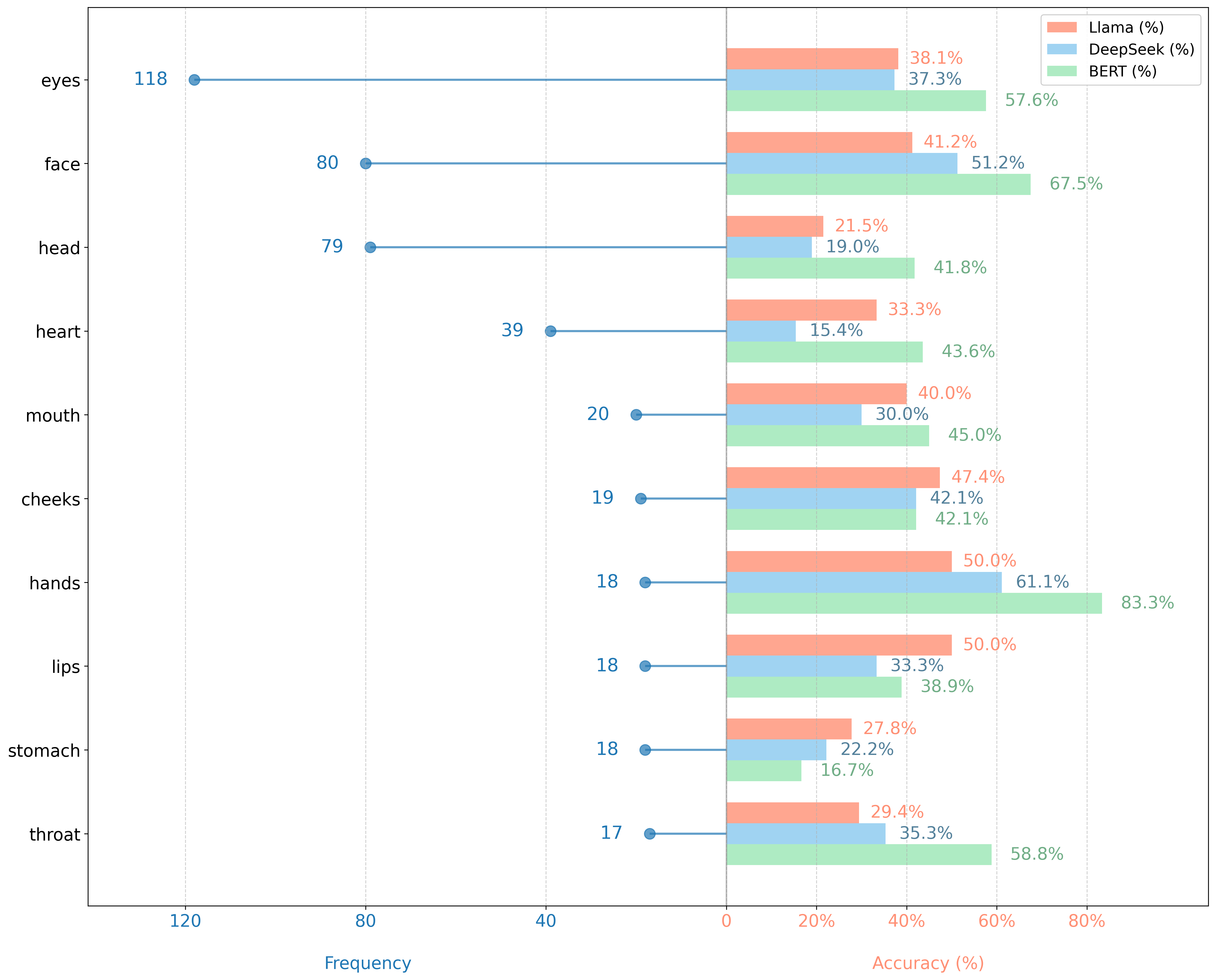}
    \caption{Embodied emotion classification of top 10 frequent body parts with frequency (left) and accuracy (right) comparing zeroshot with Llama, DeepSeek and finetuned BERT.}
    \label{fig:llama_frequent_body_part_accuracy}
\end{figure*}

\begin{figure*}[h]
    \centering
    \includegraphics[width=0.6\linewidth]{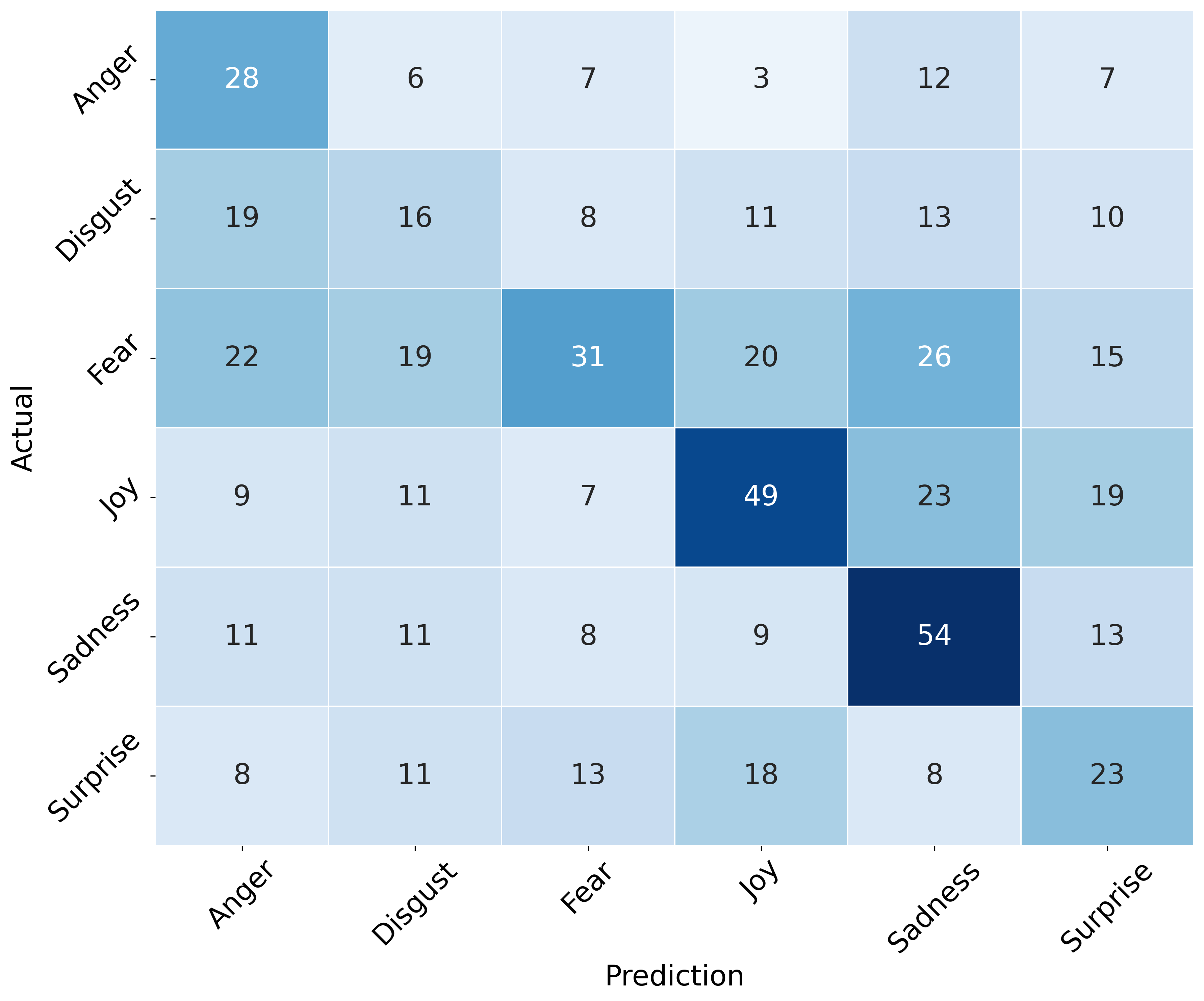}
    \caption{Confusion matrix of Llama predictions.}
    \label{fig:llama_heatmap}
\end{figure*}

\begin{figure*}[h]
    \centering
    \includegraphics[width=0.6\linewidth]{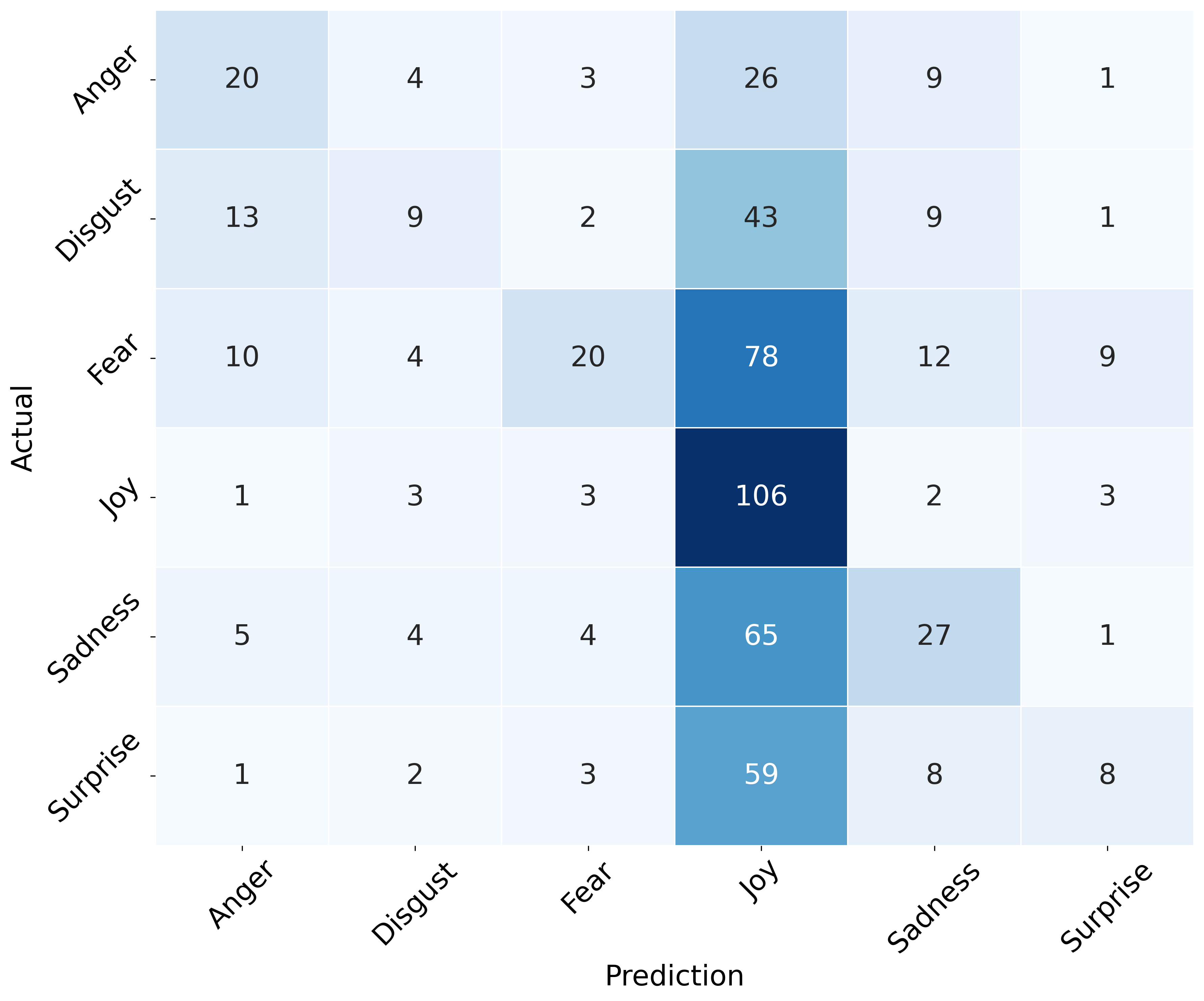}
    \caption{Confusion matrix of DeepSeek predictions.}
    \label{fig:ds_heatmap}
\end{figure*}

Figure~\ref{fig:bodypart_distributions} -~\ref{fig:ds_heatmap} present a more detailed analysis of the CHEER-Ekman dataset and evaluation of models' performance. Figure~\ref{fig:bodypart_distributions} illustrates the frequency of the top 10 body parts associated with each emotion, where the size of the bubble reflects the co-occurrence of the body part and emotion pair. Notably, the body parts \textit{face}, \textit{eye}, \textit{head}, \textit{hand}, and \textit{throat} appear consistently across the top 10 body parts across all emotions, with the highest frequency observed in \textit{face}, \textit{eye}, and \textit{head}. 

Figure~\ref{fig:llama_frequent_body_part_accuracy} illustrates the classification performance of the 10 most frequent body parts, with their frequency and corresponding accuracy across the three language models: Llama-3.1-8B, DeepSeek-R1-8B, and fine-tuned BERT. As expected, the fine-tuned BERT model consistently outperforms both Llama and DeepSeek for the most frequent body parts. Generally, the fine-tuned BERT model outperforms both zero-shot Llama and Deepseek, achieving an average accuracy increase of 11.7 over Llama and 14.8 over DeepSeek across the top 10 frequent body parts.

Figure~\ref{fig:llama_heatmap} and~\ref{fig:ds_heatmap} present confusion matrices comparing the models' predicted emotions against the ground truth emotion for Llama and DeepSeek, respectively. When comparing the two figures, a notable pattern emerges. In Figure~\ref{fig:llama_heatmap}, strong activations across the diagonal indicate Llama's attempt to predict emotions accurately without bias towards any one emotion. In Figure~\ref{fig:ds_heatmap}, however, we observe a prominent concentration in \textit{Joy} in the DeepSeek model.

\begin{table*}[h]
    \renewcommand{\arraystretch}{1.3}
    \setlength{\tabcolsep}{5pt}
    \centering
    \small
    \begin{tabularx}{\textwidth}{p{2.5cm} X}
        \toprule
        \textbf{Task} & \textbf{Prompt Template} \\
        \midrule
        Base & Please determine if a body part is involved in any embodied emotion. Specifically, a body part is involved in some embodied emotion if both conditions below are satisfied: \newline
        1) The physical movement or physiological arousal involving the body part is evoked by emotion. \newline
        2) The physical movement, if there is any, has no other purpose than emotion expression. \newline
        Answer "True" if the body part is involved in any embodied emotion, and "False" otherwise. \newline \newline
        Preceding Context: \texttt{<preceed|>} \newline
        Sentence: \texttt{<sentence|>} \newline
        Body part: \texttt{<bdypart|>} \newline
        Answer: \\
        \midrule
        Simple & Decide if a body part is used purely to express emotion. Ask: \newline
        - Did emotion cause the body part’s movement/response? \newline
        - Was the movement ONLY for expressing emotion (no other reason)? \newline
        If both are true, say "True." Else, say "False." \newline \newline
        Preceding Context: \texttt{<preceed|>} \newline
        Sentence: \texttt{<sentence|>} \newline
        Body part: \texttt{<bdypart|>} \newline
        Answer: \\
        \bottomrule
    \end{tabularx}
    \caption{Zero-shot templates for different tasks.}
    \label{tab:zs-fs-prompts}
\end{table*}

\begin{table*}[h]
    \renewcommand{\arraystretch}{1.3}
    \setlength{\tabcolsep}{5pt}
    \centering
    \small
    \begin{tabularx}{\textwidth}{p{2.5cm} X}
        \toprule
        \textbf{Setting} & \textbf{Prompt Template} \\
        \midrule
        2-Step & Please determine if a body part is involved in any embodied emotion. \newline \newline
        First, answer Condition 1: Is the body part's movement/arousal caused by emotion? \newline
        Then, answer Condition 2: Does the movement lack non-emotional purposes?  \newline \newline
        If both of those conditions are true, answer "True." Otherwise, answer "False." Please reason step-by-step for your answer.\newline
        Here is the question: \newline \newline
        Preceding Context: \texttt{<preceed|>} \newline
        Sentence: \texttt{<sentence|>} \newline
        Body part: \texttt{<bdypart|>} \newline \\
        \midrule
        3-Step & Please determine if a body part is involved in any embodied emotion. Specifically, a body part is involved in some embodied emotion if both conditions below are satisfied:
        Before answering, reasoning step-by-step \newline \newline
        1. Identify the body part mentioned. \newline
        2. Check if emotion directly caused its movement/arousal. \newline
        3. Verify if the movement has no functional purpose.  \newline \newline
        Only if all of the above are true, answer "True." Otherwise, answer "False." \newline
        Here is the question: \newline \newline
        Preceding Context: \texttt{<preceed|>} \newline
        Sentence: \texttt{<sentence|>} \newline
        Body part: \texttt{<bdypart|>} \newline \\
        \midrule
        2-Step Simple & Decide if a body part is used purely to express emotion. Ask: \newline \newline
        - Did emotion cause the body part’s movement/response? \newline
        - Was the movement ONLY for expressing emotion (no other reason)?  \newline
        If both are true, say "True." Else, say "False." Before answering, give your reasoning step-by-step.\newline \newline
        Preceding Context: \texttt{<preceed|>} \newline
        Sentence: \texttt{<sentence|>} \newline
        Body part: \texttt{<bdypart|>} \newline \\
        \bottomrule
    \end{tabularx}
    \caption{Chain of Thought (CoT) prompt templates for different settings.}
    \label{tab:cot-prompts}
\end{table*}

\begin{table*}[h]
    \renewcommand{\arraystretch}{1.3}
    \setlength{\tabcolsep}{5pt}
    \centering
    \small
    \begin{tabularx}{\textwidth}{X}
        \toprule
        \textbf{Prompt Template} \\
        \midrule
        Classify the emotion expressed by the body part in a sentence into one of six categories: "Joy", "Sadness", "Anger", "Fear", "Surprise", or "Disgust". \newline \newline
        Preceding Context: \texttt{<preceed|>} \newline
        Sentence: \texttt{<sentence|>} \newline
        Body part: \texttt{<bdypart|>} \newline
        Answer:\\
        \bottomrule
    \end{tabularx}
    \caption{Emotion classification prompt template.}
    \label{tab:emo-prompts}
\end{table*}

\begin{table*}[h]
    \renewcommand{\arraystretch}{1.3}
    \setlength{\tabcolsep}{5pt}
    \centering
    \small
    \begin{tabularx}{\textwidth}{X}
        \toprule
        \textbf{Prompt Template} \\
        \midrule
        You are an expert annotator specializing in emotion recognition. Rank the following examples based on how much \texttt{<emo|>} the specified body part exudes in the text.\\ \\

        Instructions: \newline
            - Use only the Preceding Text for context. \newline
            - Identify which example conveys the MOST \texttt{<emo|>} and which conveys the LEAST \texttt{<emo|>} based on the body part mentioned. \newline
            - Do not repeat the text. Only provide the Example numbers in the specified format. \\ \\

        Example: \texttt{<textid|>} \newline
            Preceding Context: \texttt{<preceed|>} \newline
            Sentence: \texttt{<sentence|>} \newline
            Body part: \texttt{<bdypart|>} \newline

        Example: \texttt{<textid|>} \newline
            Preceding Context: \texttt{<preceed|>} \newline
            Sentence: \texttt{<sentence|>} \newline
            Body part: \texttt{<bdypart|>} \newline

        Example: \texttt{<textid|>} \newline
            Preceding Context: \texttt{<preceed|>} \newline
            Sentence: \texttt{<sentence|>} \newline
            Body part: \texttt{<bdypart|>} \newline

        Example: \texttt{<textid|>} \newline
            Preceding Context: \texttt{<preceed|>} \newline
            Sentence: \texttt{<sentence|>} \newline
            Body part: \texttt{<bdypart|>} \newline

        Format your response as: \newline
        Most \texttt{<emo|>} Example: \newline
        Least \texttt{<emo|>} Example: \newline \\
        \bottomrule
    \end{tabularx}
    \caption{BWS-Emotion classification prompt template.}
    \label{tab:emo-bws-prompts}
\end{table*}

\end{document}